\documentclass[11pt,twoside]{article}
\usepackage[utf8x]{inputenc}
\usepackage[mathscr]{euscript}
\PrerenderUnicode{Ö}
\PrerenderUnicode{ö}
\PrerenderUnicode{ü}
\usepackage{ap-article}

\usepackage[numbers]{natbib}
\renewcommand{\cite}{\citep}

\newcommand{\titleName}{Sparsity-driven weighted ensemble classifier}

\usepackage{amsmath,amsfonts,amssymb}
\usepackage{adjustbox}
\usepackage{booktabs}
\usepackage{url}
\usepackage{algorithm}
\usepackage{algorithmicx}
\usepackage{algpseudocode}

\usepackage{multirow}

\usepackage[colorinlistoftodos,prependcaption,textsize=tiny]{todonotes}

\newcommand{\matr}[1]{\mathbf{#1}}

\usepackage[flushleft]{threeparttable}
\def\labart{yourLabel}      
\shortauthors{A.Özgür F.Nar H.Erdem}
\shorttitle{\titleName}
\title{%
\titleName
}
\author{%
Atilla Özgür\,\up{1}\,
Fatih Nar\,\up{2}
Hamit Erdem\,\up{3}
}
\addresses{%
\up{1}
Logistics Engineering, Jacobs University,\\
Campus Ring 1 28759 
Bremen, Germany \\ 
\vspace*{0.04truein}
E-mail: a.oezguer@jacobs-university.de
\\ \vspace*{0.05truein}
\up{2}
Computer Engineering, Konya Food and Agriculture University,\\
Dede Korkut Mah. Beyşehir Cad. No:9 ,\\ 
Meram / Konya / Turkey \\ 
\vspace*{0.04truein}
E-mail: fatih.nar@gidatarim.edu.tr
\\ \vspace*{0.05truein}
\up{3}
Electrical Engineering, Başkent University,\\
Bağlıca Kampüsü Fatih Sultan Mahallesi 
Eskişehir Yolu 18. km ,\\
Ankara 06790, Turkey \\ 
\vspace*{0.04truein}
E-mail: herdem@baskent.edu.tr
}
\begin{document}
\label{\labart-FirstPage}

\maketitle
\begin{abstract}

In this study, a novel sparsity-driven weighted ensemble classifier (SDWEC) that improves classification accuracy and minimizes the number of classifiers is proposed. 
Using pre-trained classifiers, an ensemble in which base classifiers votes according to assigned weights is formed.
These assigned weights directly affect classifier accuracy.
In the proposed method, ensemble weights finding problem is modeled as a cost function with the following terms: (a) a data fidelity term aiming to decrease misclassification rate, (b) a sparsity term aiming to decrease the number of classifiers, and (c) a non-negativity constraint on the weights of the classifiers. 
As the proposed cost function is non-convex thus hard to solve, convex relaxation techniques and novel approximations are employed to obtain a numerically efficient solution. 
Sparsity term of cost function allows trade-off between accuracy and testing time when needed. 
The efficiency of  SDWEC was tested on 11 datasets  and  compared  with  the state-of-the art classifier ensemble methods. 
The results show that SDWEC provides better or similar accuracy levels using fewer classifiers and reduces testing time for ensemble.


\end{abstract}

\medskip
\keywords{
   Machine Learning, Ensemble, Convex Relaxation, Classification, Classifier Ensembles 
}

\vspace*{7pt}\textlineskip



\begin{multicols}{2}

\section{Introduction}
\label{section-Introduction}

The accuracy of classification can be improved by using more than one classifier. 
This process is known by different names in different domains such as classifier fusion, classifier ensemble, classifier combination, mixture of experts, committees of neural networks, or voting pool of classifiers etc \cite{Kuncheva2001Decision}.

Ensembles can be categorized as weak or strong depending on the used classifier type \cite{Freund1995desicion}.
The weak classifiers use machine learning algorithms with fast training times and lower classification accuracy.
Due to fast training times, weak classifier ensembles contain high number of classifiers, such as 50--200 classifiers.
On the other hand, strong classifiers have slow training times and high generalization accuracy individually.
Due to slow training times, strong classifier ensembles contain as low as 3--7 classifiers.

Although using more classifiers increases generalization performance of ensemble classifier, this degrades after a while. 
To put it in another way, similar classifiers do not contribute to overall accuracy very much.  
This deficiency can be removed by increasing the classifier diversity \cite{Kuncheva2001Decision,Zhang2014Weighted,Kuncheva2003Measures}.  
Therefore, finding new diversity measurements \cite{Krawczyk2014Diversity} and improving  existing ones \cite{Kuncheva2003Measures} are an ongoing research effort in ensemble studies.

Research in the ensembles can be categorized into two groups according to their construction methods:
(a) Combining pre-trained classifiers.
(b) Constructing classifiers and ensemble together.

Methods in the first group (a) are the easiest to understand and the mainly used methods to create ensembles.
The classifiers are trained using training set and combined in an ensemble.
The simplest method to ensemble classifiers is majority (plurality) voting.
In the majority voting method, every classifier in an ensemble gets a single vote for result.
The output is the most voted result.
A well-known approach that uses majority voting in its decision stage is Bootstrap aggregating algorithm (Bagging)  \cite{Breiman1996Bagging}.
Bagging trains weak classifiers from same dataset using uniform sampling with replacement, then classifiers are combined using simple majority voting \cite{Kuncheva2005Combining}.
Instead of using a single vote for every classifier, weighted voting might be used \cite{Kuncheva2005Combining}. 
Standard Weighted majority voting (WMV) algorithm \cite{Kuncheva2005Combining} uses accuracy of individual classifiers for finding weights.
Classifiers that have better accuracies in training step get higher weights for their votes, and become more effective in voting. 

Kuncheva and Rodriguez \cite{Kuncheva2014weighted} proposed a probabilistic framework for classifier ensembles.
This framework shows relationships between four combiners: majority voting, weighted voting, recall voting, and naive bayes voting.
According to the experiments of Kuncheva and Rodr{\'i}guez \cite{Kuncheva2014weighted} on 73 benchmark datasets, there is no definite best combiner among those four.
These results conform to ``no free lunch theorem'' \cite{,Wolpert2002Supervised}.
No universal classifier exists that is suitable for every problem.
Numerous other methods has been proposed for finding weights to combine pre-trained classifiers, Table~\ref{table-Related}.
Methods in Table~\ref{table-Related} are also summarized in Section~\ref{section-Related Works}.
Similar to approaches in Table~\ref{table-Related}, main focus of this study is to present a new approach for finding weights in an ensemble that uses pre-trained classifiers using convex optimization techniques.

In the second categorization (b), ensemble construction and classifier construction affect each other.
Adaboost \cite{Freund1999short} is a well known example for this categorization that trains weak classifiers iteratively and adds them to ensemble.
Different from bagging, subset creation is not randomized in boosting.
At each iteration, subsets are obtained using results of previous iterations.
That is miss-classified data in previous subsets are more likely included.
In classifier ensemble, standard weighted majority voting is used.


Gurram and Kwon \cite{Gurram2013Sparse} used similar approach to classify remote sensing images.
Randomly selected features were used to train weak SVM classifiers.
Optimization process of training and combination of these classifiers were done together.
Lee et al. \cite{Lee2012new} combined neural network weak classifiers in their ensemble.
Genetic algorithms were used for finding weights for neural network neurons and increase diversity among neural networks.
Then, these diverse neural networks were combined using negative correlation rule.
Neural networks were trained and combined in one step.
Tian and Fend \cite{Tian2014Adaptive} proposed an approach that combines feature sub-selection and ensemble optimization.
They proposed three-term cost function: a classification accuracy term, a diversity term and a feature size term.
They solved this ensemble cost function using population based heuristics optimization.
Zhang et al. \cite{Zhang2015Kernel} used Kernel sparse representation based classifiers for ensemble in face recognition domain.
Features were projected to higher dimensions using kernels, then sparse representation of these features were found using optimization techniques.
Similarly, Kim et al. \cite{Kim2015Ensemble} proposed ensemble approach for biological data.
Their approach were similar to boosting but they selected sparse features in their weak classifiers.
Özgür and Erdem \cite{Oezguer2018} used Genetic algorithms to select features and find weights for classifier ensemble in their study.
They combined different strong classifiers and experimented on NSL-KDD dataset.

\subsection{Related works: ensembles that combine pre-trained classifiers}
\label{section-Related Works}

\begin{table*}[t]
\centering
\caption{Ensemble weights finding studies that use pre-trained classifiers}
\label{table-Related}

\begin{adjustbox}{width=\linewidth,keepaspectratio}
\begin{threeparttable}

\begin{tabular}{@{}p{0.25\linewidth}p{0.1\linewidth}p{0.25\linewidth}p{0.25\linewidth}p{0.05\linewidth}p{0.05\linewidth}p{0.2\linewidth}p{0.2\linewidth}p{0.05\linewidth}@{}}
Study & Year  & Classifiers & Method & Size & Sparse & Cost Function & Regularizer & Notes \\

\toprule

Sylvester and Chawla \cite{Sylvester2006Evolutionary} & 2006  & 12 Different Classifiers & Genetic Algorithms & 120 & No &  No Information & No Information  & \\

Li and Zhou \cite{Li2009Selective} & 2009  & Decision Tree & Quadratic Programming & 100 & Yes & Hinge Loss & $L_1$  & \\

Kim et al. \cite{Kim2011weight} & 2011  & Decision Tree & Matrix Decomposition & 64 & No & Indicator Loss & No Regularization  & \\

Mao et al. \cite{Mao2011Greedy} & 2011  & Decision Tree, SVM\tnote{1} & Matching Pursuit & 100 & Yes & Sign Loss & No Regularization  \\

Zhang and Zhou \cite{Zhang2011Sparse} & 2011  & KNN\tnote{2} & Linear Programming & 100 & Yes & Hinge Loss & $L_1$  &  \\

Goldberg and Eckstein \cite{Goldberg2012Sparse} & 2012   & No Information  & Linear Programming & NI & Yes & Indicator Loss & $L_0$  & \textbf{\tnote{a}} \\

Santos et al. \cite{Santos2012Combiner} & 2012  & SVM\tnote{1}, MLP\tnote{3} & Genetic Algorithms & 6 & No & No Cost Function & No Regularization  &  \\

Yin et al. \cite{Yin2012Classifier} & 2012  & Neural Networks  & Genetic Algorithms & 100 & Yes & Square Loss & $L_1$  & \textbf{\tnote{b}}\\

Meng and Kwok \cite{Meng2013Enhancing} & 2013 & Decision Tree, SVM\tnote{1}, KNN\tnote{2} & Domain Heuristic & 3 & No & No Cost Function & No Regularization  &  \\

Tinoco et al. \cite{Tinoco2013Ensemble} & 2013  & SVM\tnote{1}, MLP\tnote{3} & Linear Programming & 6 & Yes & Hinge Loss  & $L_1$  & \textbf{\tnote{d}}  \\

Hautamaki et al. \cite{Hautamaeki2013Sparse} & 2013  & Logistic Regression & Nelder–Mead & 12 & Yes & cross-entropy \cite{Bishop2006Pattern} & $L_1, L_2, L_1 + L_2$  & \textbf{\tnote{c}} \\

Şen and Erdoğan \cite{Sen2013Linear} & 2013  & 13 Different Classifiers  & Convex Opt.  & 130 & Yes & Hinge Loss & $L_1$ , Group Sparsity  & \\

Mao et al. \cite{Mao2013Weighted} & 2013  & Decision Tree & Singular Value Decomposition & 10 & No & Absolute Loss & No Regularization  & \\

Yin et al. \cite{Yin2014novel} & 2014  & Neural Networks  & Genetic Algorithms & 100 & Yes & Square Loss & $L_1$  &  \textbf{\tnote{e}} \\
Yin et al.\cite{Yin2014Convex} & 2014  & Neural Networks  & Quadratic Programming & 100 & Yes & Square Loss & $L_1$  & \textbf{\tnote{f}} \\

Zhang et al. \cite{Zhang2014Weighted} & 2014  & 5 Different Classifiers & Differential Evolution & 5 & No & No Cost Function & No Regularization  \\

Mao et al. \cite{Mao2015Weighted} & 2015  & Decision Tree & Quadratic Form & 200 & No & Square Loss & $L_1$   & \\

\bottomrule
\end{tabular}

\begin{tablenotes}
\setlength{\columnsep}{0.8cm}
\setlength{\multicolsep}{0cm}
  \begin{multicols}{3}
        \item[1] SVM Support Vector Machines.
        \item[2] KNN K-Nearest Neighbor 
        \item[3] MLP Multi Layer Perceptron.
        \item[a] No experimental results.
        \item[b] Diversity Term Yule's Q Statistic is used 
        \item[c] Improved version of \cite{Santos2012Combiner}
        \item[d] 3 regularizers are compared 
        \item[e] Journal version of \cite{Yin2012Classifier} 
        \item[f] Convex Quadratic model of \cite{Yin2014novel} and \cite{Yin2012Classifier} 
  \end{multicols}
\end{tablenotes}

\end{threeparttable}

\end{adjustbox}

\end{table*}



Focus of this study is to combine pre-trained classifiers so that combined accuracy of the ensemble is better than individual classifiers.
This study aims to accomplish this objective in a sparse manner so that not all of the classifiers are used in ensemble; therefore, weak decision tree classifiers are used in the experiments.
Although some of the other sparse approaches \cite{Gurram2013Sparse,Zhang2015Kernel,Kim2015Ensemble,Shukla2015novel} are mentioned before, in this section, only ensemble classifiers that proposed methods to find weights for base classifiers are investigated.

Sylvester and Chawla \cite{Sylvester2006Evolutionary} proposed differential evolution to find suitable weights for ensemble base classifiers.
Similar to most heuristic solution techniques, they did not explicitly define cost function, but use classification accuracy for fitness function.
ID3 decision trees, J48 decision trees (C4.5), JRIP rule learner (Ripper), Naive Bayes, NBTree (Naive Bayes trees), One Rule, logistic model trees, logistic regression, decision stumps, multi-layer perceptron (MLP), SMO (support vector machine), and 1BK (k-nearest neighbors) classifiers from Weka toolbox \cite{hall2009weka} were used in the experiments.

Li and Zhou \cite{Li2009Selective} modeled ensemble weights finding problem using cost function that consists of hinge loss and $L_1$ regularization.
This cost function were minimized using Quadratic programming.
Decision tree weak classifiers and UCI datasets were used for experiments.
A semi-supervised version was also suggested.

Zhang and Zhou \cite{Zhang2011Sparse} formulated weights finding problem using three different cost functions:
LP1 uses a cost function that consists of Hinge loss only.
LP2 uses a cost function that consists of Hinge loss and $L_1$ regularization.
LP3 allows weights to be negative.
These cost functions were minimized using linear programming.
They used K-Nearest neighbor (KNN) algorithm as base classifiers and UCI datasets in their experiments.

Kim et al. \cite{Kim2011weight} proposed an approach similar to boosting.
They considered two weight vectors, one for classifier and one for instances.
Hard to classify instances get more weight and correspondingly they affect weight vector more.
Different from boosting, their approach works with pre-trained classifiers.
Weights for ensemble was found using matrix decomposition and an iterative algorithm.
Decision tree weak classifiers and UCI datasets were used for experiments.

Mao et al. \cite{Mao2011Greedy} proposed matching pursuit algorithm to find weights for ensemble base classifiers.
Since matching pursuit is a sparse approximation algorithm \cite{Mallat1993Matching}, their approach include sparsity.
Decision Tree and SVM weak classifiers and UCI datasets were used for experiments.

Goldberg and Eckstein \cite{Goldberg2012Sparse} modeled weights finding problem with indicator loss function and $L_0$ regularization function. 
According to Goldberg and Eckstein \cite{Goldberg2012Sparse}, this problem is NP-hard in special cases.
They gave different relaxation strategies to solve this problem and gave their relaxation bounds.
Different from other studies, this study was purely theoretical.

Santos et al. \cite{Santos2012Combiner} combined MLP and SVM algorithms to classify remote sensing images.
They did not give any explicit cost function but used genetic algorithms for finding weights.
An improved version of their studies \cite{Tinoco2013Ensemble} modelled weights finding problem using hinge loss and $L_1$ regularization.
This cost function were minimized using linear programming.
In both versions, remote sensing images were classified using ensemble of MLP and SVM classifiers.

Mong and Kwok \cite{Meng2013Enhancing} combined Decision Tree(J48), K-Nearest Neighbor and SVM classifiers.
They suggest using following domain heuristic for weights of classifiers: 
"...weighted ranking (precision of false alarm $>$ recall of false alarm $>$ classification accuracy) is an appropriate and correct way to decide the weight values with high confidence in ensemble selection..." \cite{Meng2013Enhancing}.

Hautamaki et al. \cite{Hautamaeki2013Sparse} investigated using sparse ensemble in speaker verification domain.
Ensemble weights finding problem were modeled using cross-entropy loss function and three different regularization functions:  $L_1$, $L_2$, and $L_1+L_2$.
These cost functions were minimized using Nelder–Mead method.
Logistic regression classifiers were used in experiments.

Yin et al. \cite{Yin2012Classifier} modeled ensemble weights finding problem with a cost function that consists of the terms a square loss, $L_1$ regularization and a diversity based-on Yule's Q statistics.
They used neural network classifiers on 6 UCI datasets in their experiments.
In their first study \cite{Yin2012Classifier}, the proposed cost function were minimized using genetic algorithms.
In their second study \cite{Yin2014novel}, the Pascal 2008 webspam dataset were added to their experiments.
Finally, convex optimization techniques \cite{Yin2014Convex} were used to minimize the same cost function.

Sen and Erdogan \cite{Sen2013Linear} modeled ensemble weights finding problem using a cost function that consists Hinge loss and two different regularization functions, $L_1$ and group sparsity.
This cost function were minimized using convex optimization techniques.
In their experiments, 13 different classifiers were compared on 12 UCI datasets and 3 other datasets using CVX Toolbox \cite{Grant2014CVX,Grant2008Graph}.

Zhang et al. \cite{Zhang2014Weighted} proposed Differential Evolution for finding suitable weights for ensemble base classifiers.
Similar to most heuristic solution techniques, they did not explicitly define cost function, but use classification accuracy for fitness function.
Decision Tree (J4.8), Naive Bayes, Bayes Net, K-Nearest Neighbor, and ZeroR classifiers from Weka toolbox \cite{hall2009weka} were used in the experiments.

Mao et al. \cite{Mao2013Weighted} modeled ensemble weights finding problem using a cost function that consists of absolute loss only.
Proposed cost function was minimized using  0--1 matrix decomposition.
In a later study \cite{Mao2015Weighted}, Mao et al.  proposed a cost function that consists of square loss and $L_1$ regularization function.
This cost function was minimized using quadratic form approximation. 
Both studies used decision tree weak classifiers and UCI datasets in experiments.

As can be seen from Table~\ref{table-Related}, numerous approaches exist for finding weights in ensemble classification.
Inspired from studies of \cite{Zhang2011Sparse,Mao2013Weighted,Mao2015Weighted,Nar2016Sparsity,Oezguer2018},  sparsity-driven weighted ensemble classifier (SDWEC) has been proposed.
SDWEC can use both strong classifiers or weak classifiers for classifier ensemble. 
In this study, decision tree as a weak classifier is used as the base classifier since choosing fewer number of classifiers among large number of weak classifiers leads to high accuracy with shorter testing time.
Proposed cost function consists of the following terms: (1) a data fidelity term with sign function aiming to decrease misclassification rate, (2) $ L_1$-norm sparsity term aiming to decrease the number of classifiers, and (3) a non-negativity constraint on the weights of the classifiers.
Cost function proposed in SDWEC is hard to solve since it is non-convex and non-differentiable; thus, (a) the sign operation is convex relaxed using a novel approximation, (b) the non-differentiable $L_1$-norm sparsity term and the non-negativity constraint are approximated using \textit{log-sum-exp} and Taylor series.
Contributions of SDWEC can be summarized as follows:

\begin{enumerate}
    \item A new cost function is proposed for ensemble weights finding problem.
    \item This cost function is minimized using novel convex relaxation and approximation techniques for sign function and absolute value function. 
    \item SDWEC provides similar or better classification accuracy, while minimizing the number of classifiers used in the ensemble.
    \item According to sparsity level of SDWEC, number of classifiers used in the ensemble decreases; thus, the testing time for whole ensemble decreases.
    \item The sparsity level of SDWEC allows trade-off between testing accuracy and testing time when needed.
    \item Computational Complexity of SDWEC is analyzed theoretically and experimentally, which is linear in number of data rows, number of classifiers and number of algorithm iterations.
\end{enumerate}



\section{Sparsity-driven weighted ensemble classifier}
\label{section-MaterialAndMethods}

An ensemble consists of $l$ number of classifiers which are trained using training dataset.
We aim to increase ensemble accuracy on test dataset by finding suitable weights for classifiers using validation dataset.
Ensemble weights finding problem is modeled with the following matrix equation.

$
\underbrace{
sgn(
\begin{bmatrix}
-1 & -1 &  \dots &  +1 \\ 
+1 & -1 &  \dots &  -1 \\ 
\vdots &  \vdots & \vdots & \vdots \\ 
-1 & \dots  &  \dots & +1 \\ 
+1 & \dots  &  \dots & -1 \\ 
\end{bmatrix}}_{\matr{H}_{m x l}}
\underbrace{
\begin{bmatrix}
w_1\\ 
w_2\\ 
\vdots \\ 
w_{l-1}\\ 
w_l\\ 
\end{bmatrix}}_{w_{l x 1}}
)
 \approx
\underbrace{
\begin{bmatrix}
y_1\\ 
y_2\\ 
 \vdots \\ 
y_{m-1}\\ 
y_m\\ 
\end{bmatrix}}_{y_{m x 1}}$

\begin{tabular}{p{0.10\columnwidth}p{0.75\columnwidth}}
$\matr{H}$  & classifiers results $\{-1,1\}^{m \text{\tiny{x}} l}$ \\
$m$  & number of samples in the validation dataset \\
$l$  & number of individual classifiers \\
$w$  & classifier weights \\
$y$  & true labels $\{-1,1\}^{m\text{\tiny{x}}1}$ for the validation dataset \\  
\end{tabular}


In this matrix equation, classifiers predictions are weighted so that obtained prediction for each data row becomes approximately equal to expected results.
Matrix $\matr{H}$ consists of $l$ classifier predictions for $m$ data rows that are drawn from validation dataset.
Vector $y$ contains the labels for the validation dataset.
Our aim is to find suitable weights for $w$ in a sparse manner while preserving condition of  $ sgn(\matr{H}w) \approx y$ (sign function). For this model, the following cost function is proposed:

\begin{align} 
\label{eq-cost-function_constrained}
\begin{split}
J(w) &=  \frac{\lambda}{m} \sum_{s=1}^{m} (sgn(H_s w)-y_s)^2 + \frac{1}{l}||w||_1^1 \,  \\
     & \textrm{  subject to } w  \geq 0
\end{split}
\end{align}


\begin{tabular}{p{0.10\columnwidth}p{0.75\columnwidth}}
$\lambda$   &  data fidelity coefficient ($\lambda > 0$)  \\
$H_s$  &  $s_{th}$ row vector of matrix $\matr{H}$   \\
$y_s$  &  $s_{th}$ label for vector $y$ \\  
\end{tabular}


In equation~\ref{eq-cost-function_constrained}, the first term acts as a data fidelity term and minimizes the difference between true labels and ensemble predictions. 
Base classifiers of ensemble give binary predictions ($-1$ or $1$) and these predictions are multiplied with weights through sign function which leads to $\{-1, 0, 1\}$ as an ensemble result.
To make this term independent from data size, it is divided to $m$ (number of data rows).

The second term is a sparsity term \cite{Bach2012Optimization} that forces weights to be sparse \cite{Nar2016Sparsity}; therefore, minimum number of classifiers are utilized. 
In sparsity term, any $L_p$-norm ($0 \le p \le 1$)  can be used.
Weights become more sparse as $p$ gets closer to $0$.
However, when ($0 \le p < 1$), sparsity term becomes non-convex and thus the problem becomes harder to solve.
When $p$ is $0$ ($L_0$-norm) then problem becomes NP-hard \cite{Ge2011note}.
Here, $L_1$-norm is used as a convex relaxation of $L_p$-norm \cite{Bach2012Optimization,Tropp2006Just}.
Similar to the data fidelity term, this term is also normalized with division by $l$ (number of individual classifiers).

The third term is used as a non-negativity constraint.
Since base binary classifiers use values of $-1$ and $1$ for class labels, negative weights change sign of prediction; thus they change class label of prediction.
To avoid this problem, the constraint term is added to force weights to be non-negative.

Using the $|x|=max(-x,x)$ as the definition of $L_1$-norm and using the penalty method \cite{Bertsekas2016nonlinear} for transforming the constraint in the equation~\ref{eq-cost-function_constrained} to a penalty term (i.e. $w \geq 0 \rightarrow max(-w_r, 0), 1 \leq r \leq l$), below unconstrained cost function is obtained:

\begin{align} 
\begin{split}
\label{eq-cost-function-transformed}
     J^{(n)}(w)   &= \frac{\lambda}{m} \sum_{s=1}^{m} (sgn (H_s w)-y_s)^2 \\
            &+\frac{1}{l}\sum_{r=1}^l max(-w_r,w_r) \\
            &+ \frac{\beta^{(n)}}{l} \sum_{r=1}^l  max(-w_r,0)
\end{split}
\end{align} 

\begin{figure}[H]
\centering
 
        \includegraphics[width=0.90\columnwidth,keepaspectratio]{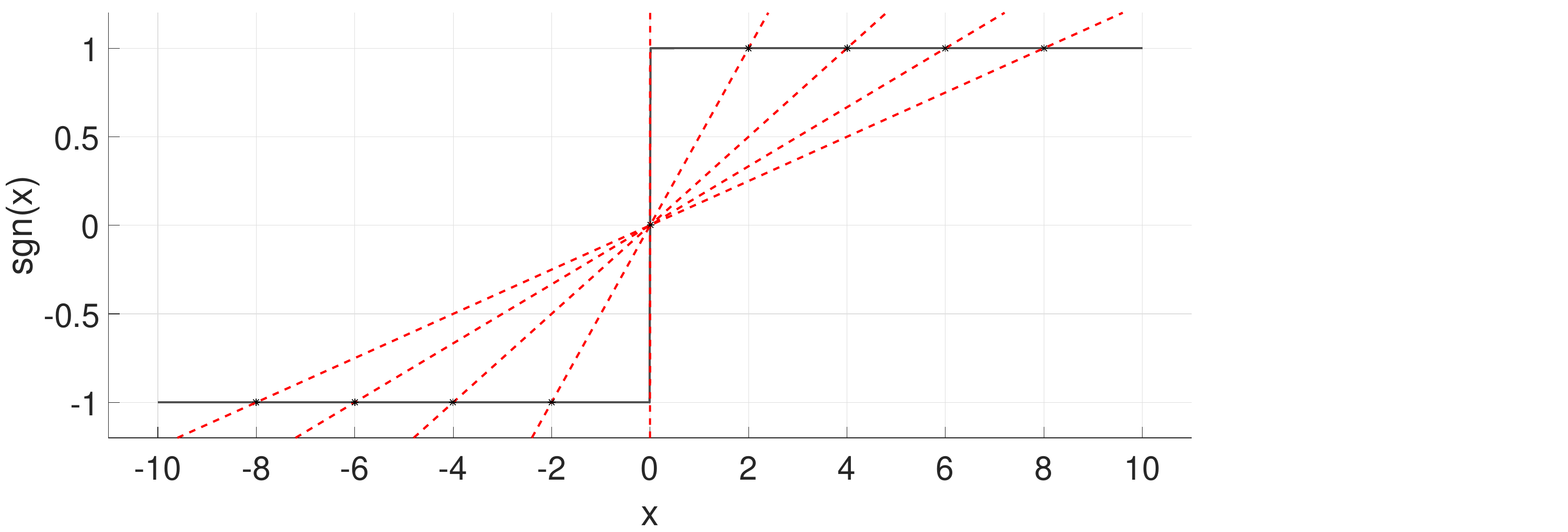}
        \vspace*{-3 mm}
        \caption[Sign function approximation using equation~\ref{eq-approximation-sgn}]{Sign function approximation using equation~\ref{eq-approximation-sgn}. Dotted Lines are approximations using Equation~\ref{eq-approximation-sgn} at various points.}
        \label{ChapterEnsembleSD-signum-approximation-plot}
        
\end{figure}

In equation~\ref{eq-cost-function-transformed}, $n$ is the iteration number since constrained cost function in equation~\ref{eq-cost-function_constrained} is converted into series of unconstrained problems using penalty method. 
Due to employed penalty method approach, the constraint $ w  \geq 0 $ is better satisfied as the penalty coefficient $\beta^{(n)}$ is increased in each iteration where $\beta^{(1)} > 0$ in the first iteration.
Equation~\ref{eq-cost-function-transformed} is a non-convex function, since \textit{sgn} function creates jumps on cost function surface.
In addition, \textit{max} function is non-differentiable.
Functions \textit{max} and \textit{sgn} in Equation~\ref{eq-cost-function-transformed} are hard to minimize.
Therefore, we propose a novel convex relaxation for \textit{sgn} as given in equation~\ref{eq-approximation-sgn}.
Figure~\ref{ChapterEnsembleSD-signum-approximation-plot} shows approximation of sign function using Equation~\ref{eq-approximation-sgn}.

\begin{align} 
\label{eq-approximation-sgn}
& sgn (H_s w) \approx \frac{H_s w}{|H_s \hat{w}|+\epsilon}  = S_s H_s w 
\end{align} 
where 
\begin{align} 
\label{eq-Ss}
 & S_s = (|H_s \hat{w}|+\epsilon)^{-1}
\end{align} 


In this equation, $\epsilon$ is a small positive constant.
We also introduce a new constant $\hat{w}$ as a proxy for $w$. 
Therefore,  $ S_s = (|H_s \hat{w}|+\epsilon)^{-1}$ is also a constant.
However, this \textit{sgn}  approximation is only accurate around introduced \textit{constant $\hat{w}$}.
Therefore, the approximated cost function needs to be solved around $\hat{w}$.
Additionally, \textit{max}  function is approximated with \textit{log-sum-exp} \cite{Boyd2004Convex} as follows:

\begin{align} 
\label{eq-approximation-soft-max-normal}
& max(-w_r,w_r)  \approx \frac{1}{\gamma} log(e^{-\gamma w_r} + e^{\gamma w_r})\
\end{align}


Accuracy of \textit{log-sum-exp} approximation becomes better as $\gamma$, a positive constant, increases.
In double-precision floating-point format \cite{1985IEEE}, values up to $10^{308}$ in magnitude  can be represented.
This means that $\gamma|w_r|$ should be less than $710$ where $exp(709) \approx 10^{308}$, otherwise exponential function will produce infinity ($\infty$).
At $w_r=0$, there is no danger of numerical overflow in exponential terms of a \textit{log-sum-exp} approximation; thus, large  $\gamma$ values can be used.
But as $|w_r|$ gets larger, there is a danger of numerical overflow in exponential terms of \textit{log-sum-exp} approximation, since $e^{\gamma|w_r|}$ may be out of double precision floating point upper limit.

\begin{figure*}[t]
\centering
        \includegraphics[width=0.825\textwidth]{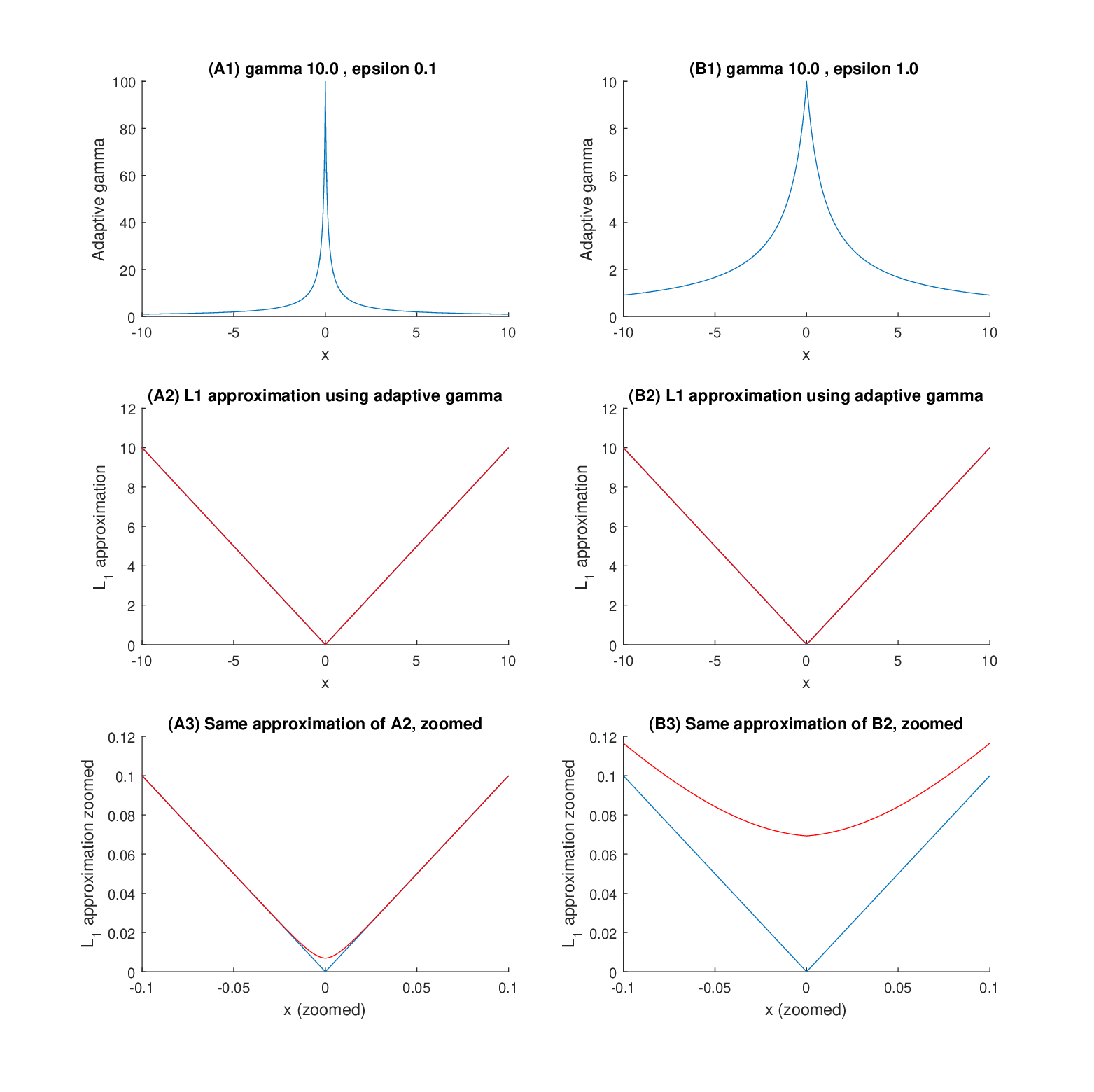}
        \vspace*{-12.5 mm} 
        \caption{Adaptive gamma ($\gamma_1$) $L_1$ Approximation with different $\epsilon$ values.}
        \label{figure-adaptive-gamma-l1-approximation}

\end{figure*}


To remedy this numerical overflow issue, a novel adaptive $\gamma$ approximation is proposed, where $\gamma_r$ is adaptive form of $\gamma$ and defined as $\gamma_r = \gamma (|\hat{w}_r| +\epsilon)^{-1}$. 
The accuracy of approximation can be improved by decreasing $\epsilon$ or increasing  $\gamma$.
Figure~\ref{figure-adaptive-gamma-l1-approximation} shows proposed adaptive $\gamma$  and resulting approximations for two different set of values $(\gamma  = 10,\epsilon = 0.1)$ and $(\gamma  = 10,\epsilon = 1)$.

Validity of the approximation can be checked by taking the limits at $-\infty$, $0$, and $+\infty$ with respect to $w_r$.
These limits are $-x$, $\frac{\epsilon\log{2}}{\lambda_r}$, and $x$ when $w_r$ goes to $-\infty,0$, and $+\infty$. 
As $|x|$ gets larger, dependency to $\gamma$ decreases; thus, proposed adaptive $\gamma$ approximation is less prone to numerical overflow compared to standard \textit{log-sum-exp} approximation.

Regularization term given in equation~\ref{eq-slow-step-regularization} is added to the unconstrained cost function (equation~\ref{eq-cost-function-transformed}) since approximated cost function needs to be solved around $\hat{w}$. 
This new regularization term forces solution to be around $\hat{w}$ by imposing a quadratic penalty between solution and $\hat{w}$.
Due to this new regularization term, solution in each iteration will be changed slowly; thus, this new term is called as slow-step regularization.
The main drawback of penalty method is the need to increase penalty coefficient in each iteration, theoretically up to infinity, that leads to ill-conditioning in the minimization of the cost function. 
However, increase of $\beta^{(n)}$ in each iteration is not needed since during the minimization changes in the solution will be small.
These small changes is accomplished due to employed slow-step regularization.
Therefore, penalty coefficient is used as a constant $\beta$ with a small value (i.e $\beta < 10^2$) for all iterations. 
Note that, using a small value for penalty coefficient $\beta$ leads to numerically well-posed minimization problem.

\begin{align} 
\label{eq-slow-step-regularization}
\begin{split}
    \frac{1}{l}\sum_{r=1}^l (w_r-\hat{w}_r)^2
 \end{split}
\end{align}


Application of adaptive $\gamma$ approximation leads to the following equations:

\begin{align} 
\label{eq-approximation-max1}
& max(-w_r,w_r)  \approx \frac{1}{\gamma_r} log(e^{-\gamma_r w_r} + e^{\gamma_r w_r})  &\\
\label{eq-approximation-max2}
& \beta max(-w_r,0) \approx \frac{\beta}{\gamma_r}  log(e^{-\gamma_r w_r} + 1) = P(w_r)
\end{align}


Use of slow-step regularization in equation~\ref{eq-slow-step-regularization} and \textit{log-sum-exp} approximation with adaptive $\gamma$ leads to the cost function shown in equation~\ref{eq-cost-function-final-version}.

\begin{align} 
\label{eq-cost-function-final-version}
\begin{split}
J^{(n)}(w) &= \frac{\lambda}{m} \sum_{s=1}^{m} (S_s H_s w - y_s)^2  \\
 &+ \frac{1}{l}\sum_{r=1}^l  \frac{1}{\gamma_r} log(e^{-\gamma_r w_r} + e^{\gamma_r w_r}) \\
 &+ \frac{1}{l} \sum_{r=1}^l \frac{\beta}{\gamma_r} log(e^{-\gamma_r w_r} + 1) \\
 &+ \frac{1}{l}\sum_{r=1}^l (w_r-\hat{w}_r)^2
 \end{split}
\end{align}


In order to achieve a second-order accuracy and to obtain a linear solution, after taking the derivative of the cost function, equation~\ref{eq-cost-function-final-version} is expanded as a second-order Taylor series centered on $\hat{w}_r$, leading to equation~\ref{eq-cost-function-after-taylor-new-regularization-term}. 

\begin{align} 
\label{eq-cost-function-after-taylor-new-regularization-term}
\begin{split}
J^{(n)}(w) &= \frac{\lambda}{m} \sum_{s=1}^{m} (S_s H_s w - y_s)^2  \\
&+ \frac{1}{l}\sum_{r=1}^l  (A_r + B_r w_r + C_r w_r^2)  \\
&+ \frac{1}{l}\sum_{r=1}^l (w_r-\hat{w}_r)^2
 \end{split}
\end{align}


In equation~\ref{eq-cost-function-after-taylor-new-regularization-term},
$A_r$ represents constants terms while $B_r$  and $C_r$ are the coefficients of the terms $w_r$ and $w_r^2$, respectively.
If $w_r$ values differ significantly from constant point, $\hat{w}_r$, Taylor approximation diverges from true cost function.
In proposed method, employed slow-step regularization also ensures the accuracy of Taylor approximations.

Equation~\ref{eq-cost-function-after-taylor-new-regularization-term} can be written in a matrix-vector form as follows:

\begin{align} 
\label{eq-cost-function-after-taylor-new-regularization-term-matrix-vector-form}
\begin{split}
J^{(n)}(w) &= \frac{\lambda}{m} (\matr{S}\matr{H}w-y)^{\intercal}(\matr{S}\matr{H}w-y)  \\
&+ \frac{1}{l} (v_A^{\intercal} \vec{1} + v_B^{\intercal} w + w^{\intercal} \matr{C} w) \\
&+ \frac{1}{l} (w-\hat{w})^{\intercal}(w-\hat{w})
 \end{split}
\end{align}


\begin{tabular}{p{0.10\columnwidth}p{0.75\columnwidth}}
$\matr{S}$   &  matrix form of $S_s$   \\
$\vec{1}$  &  vector of ones \\
$v_A$  & vector form of $A_r$ \\
$v_B$   &  vector form of $B_r$ \\
$\matr{C}$  & diagonal matrix form of $C_r$  \\
\end{tabular}


Equation~\ref{eq-cost-function-after-taylor-new-regularization-term-matrix-vector-form} is strictly convex (see appendix for the details) thus it has a unique global minimum. 
Therefore, to minimize $J^{(n)}(w) $ in Equation~\ref{eq-cost-function-after-taylor-new-regularization-term-matrix-vector-form}, the derivative with respect to $w$ is taken and is equalized to zero.
This leads to system of linear equations:

\begin{align} 
\label{eq-cost-function-system-of-linear-equations}
\begin{split}
\matr{M}w &= b \\
where  \\
\matr{M} &= \frac{2\lambda}{m}(\matr{S}\matr{H})^{\intercal}(\matr{S}\matr{H}) + \frac{2}{l}(\matr{C} + \matr{I}) \\
b &= \frac{2\lambda}{m}(\matr{S}\matr{H})^{\intercal}y + \frac{2 \hat{w} - v_B}{l} \\
\end{split}
\end{align}


In Equation~\ref{eq-cost-function-system-of-linear-equations}, $\matr{M}$ is dense, symmetric, real, and positive definite matrix with size of $l \times l$.


Final model is solved using algorithm~\ref{code-SDWEC-Pseudo} iteratively.
Due to the employed numerical approximations and using constant $\beta$, small negative weights may occur around zero.
Since our feasible set is $ w \ge 0$, back projection to this set is performed after solving linear system at each iteration in algorithm~\ref{code-SDWEC-Pseudo}.
This kind of back-projection to feasible domain is commonly used \cite{Pock2009algorithm}.
Additionally, small weights in ensemble do not contribute to overall accuracy; therefore, these small weights are thresholded after iterations are completed.

\begin{figure*}[t]
\centering

        \includegraphics[width=0.8\textwidth,keepaspectratio]{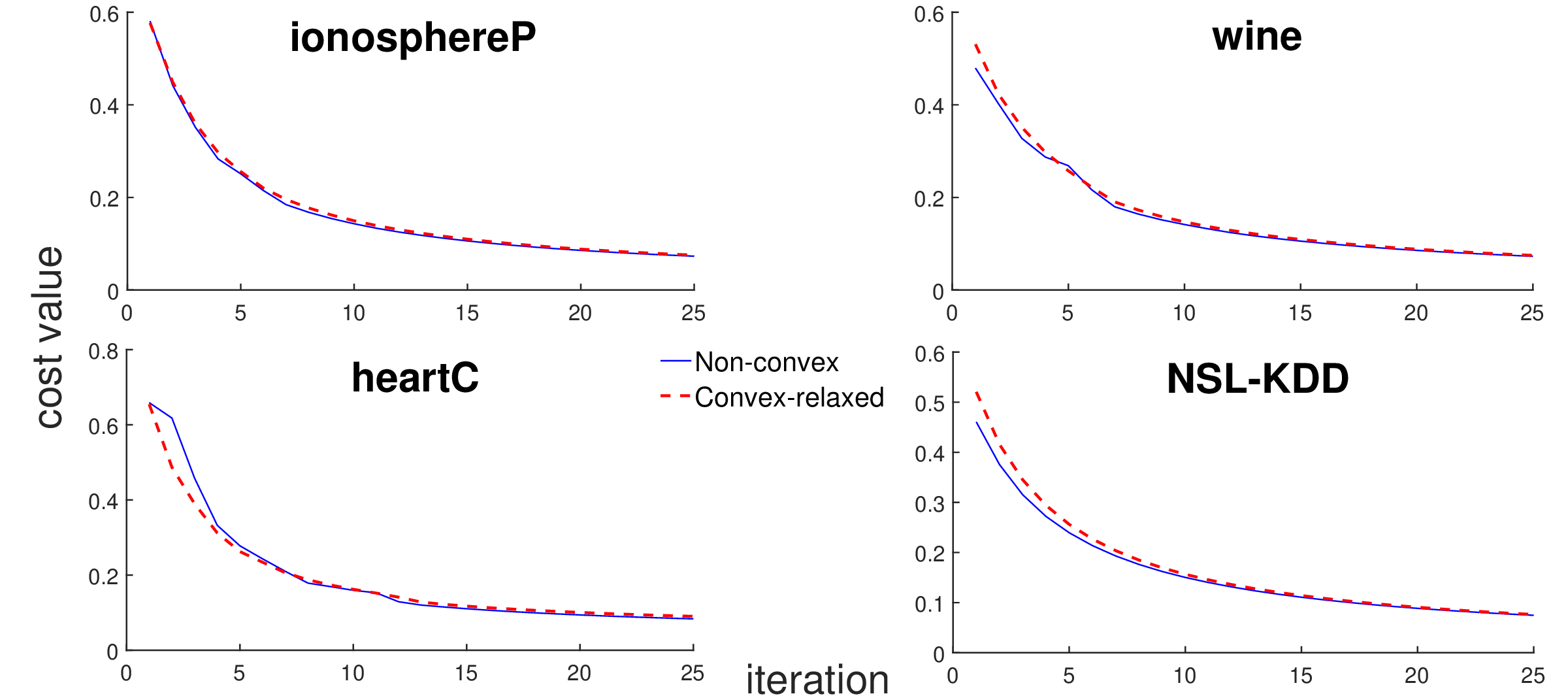}
         \vspace*{-2 mm} 
        \caption[Minimization of the cost function for 4 datasets]{Cost function minimization for 4 datasets (Non-convex equation~\ref{eq-cost-function-transformed} vs convex-relaxed equation~\ref{eq-cost-function-after-taylor-new-regularization-term-matrix-vector-form}). }
        \label{ChapterEnsembleSD-Figure-convergence-plot-Article3Plot}

\end{figure*}


\begin{algorithm}[H]
\caption{SDWEC Pseudo code}
\label{code-SDWEC-Pseudo}

\begin{algorithmic}[1]
\State $ \matr{H}, y, \lambda, \beta, \gamma, \epsilon $ are initialized 
\vspace{2mm}
\State $ w \gets \vec{1}$
\State $ m,l \gets size~of~ \matr{H}_{m x l} $
\State $ k \gets 25 $  \Comment{Maximum Iteration}
\vspace{2mm}
\For{ n = 1 \text{to} k } 
\State $\hat{w} \gets w$
\vspace{2mm}
\State    $\gamma_r \gets \frac{\gamma}{|\hat{w}| + \epsilon} $
\State construct $\matr{S}$ as diagonal form of $S_s$
\State construct $v_B$ and $\matr{C}$ 
\vspace{2mm}
\State $ \matr{M} \gets  \frac{2\lambda}{m}(\matr{S}\matr{H})^{\intercal}(\matr{S}\matr{H}) + \frac{2}{l}(\matr{C}+\matr{I})$
\State $ b \gets \frac{2\lambda}{m}(\matr{S}\matr{H})^{\intercal}y + \frac{2 \hat{w} - v_B}{l} $
\State solve $ \matr{M}w = b $ 
\vspace{2mm}
\State $ w = max(w,0)
$ \Comment{Back projection to $w\ge 0$}
\EndFor
\vspace{2mm}
\State $w_{threshold} = argmin_{w_r} (P(w_r)-10^{-3})^2$
\vspace{2mm}
\State $ w = $\bigg\{
  \begin{tabular}{cc}
  $ w $ & $if \textrm{ } w > w_{threshold}$ \\
  $ 0 $ & $ otherwise $
  \end{tabular}

\end{algorithmic}

\end{algorithm}


An example run of Algorithm~\ref{code-SDWEC-Pseudo} can be seen in 
Figure~\ref{ChapterEnsembleSD-Figure-convergence-plot-Article3Plot}, where cost values for equations~\ref{eq-cost-function-transformed} and \ref{eq-cost-function-after-taylor-new-regularization-term-matrix-vector-form} decrease steadily.
As seen in Figure~\ref{ChapterEnsembleSD-Figure-convergence-plot-Article3Plot}, the difference between non-convex cost function and its convex relaxation is minimal especially in the final iterations.
This shows that two functions converge to very similar values.
Since non-convex Equation~\ref{eq-cost-function-transformed} and convex Equation~\ref{eq-cost-function-after-taylor-new-regularization-term-matrix-vector-form} are converged to similar points, this converged points are within close proximity of the global minimum.
Non-convex Equation~\ref{eq-cost-function-transformed} and convex-relaxed Equation~\ref{eq-cost-function-after-taylor-new-regularization-term-matrix-vector-form} are close to each other due to the slow-step regularization term and employed iterative approach for numerical minimization.
These results show success of the proposed approximations.



\section{Experimental results}
\label{section-Results}
The performance of SDWEC has been tested on 11 datasets; 10 UCI datasets and NSL-KDD \cite{Oezguer2016Review}.
NSL-KDD is a popular database for intrusion detection \cite{Albayati2015Analysis,Hussain2016two,Oezguer2016Review}.
In all ensemble methods, 200 base decision tree classifiers, Classification And Regression Trees (CART) \cite{Breiman1984} are used.
SDWEC has been compared with the following algorithms : Single decision tree classifier (CART) \cite{Breiman1984}, bagging \cite{Breiman1996Bagging}, WMV \cite{Kuncheva2005Combining}, and state-of-the-art ensemble QFWEC \cite{Mao2015Weighted}.
Each dataset is divided to training (80\%), validation (10\%), and testing (10\%) datasets.
This process has been repeated 10 times for cross validation.
Mean values have been used in Table~\ref{table-comparison}.
The accuracy values for QFWEC in Table~\ref{table-comparison} are higher than original publication  \cite{Mao2015Weighted} since weights are found using validation dataset instead of training dataset, which provides better generalization.

SDWEC finds weights of ensemble for pre-trained classifiers; thus, it is divided into 3 sub tasks.
\begin{enumerate}
    \item \textit{Training base classifiers on training dataset}: 
This sub task is common for the ensemble methods which aims to combine pre-trained classifiers.
Employed pre-trained classifier can be a weak classifier or a strong classifier where generally weak classifiers are faster to train with lower accuracy and strong classifiers are slower to train with higher accuracy.
Training time of base classifiers depends on training complexity of the method which is dependent to the number of data in the training dataset ($p$), number of features ($d$), number of classes (i.e. binary, multi-label), and data characteristics.
Computational (time) complexity of base classifier training are independent from the proposed SDWEC method; thus, one can use a base classifier of his choice.
SDWEC aims to use few number of classifiers among trained $l$ base classifiers; therefore, weak decision tree classifiers (CART) \cite{Breiman1984} are used in the experiments.

    \item \textit{Finding ensemble weights on validation dataset (SDWEC training)}:  
SDWEC finds the ensemble weights of base classifiers using $y$ and $\matr{H}$.
Here, $y$ consists of true labels and $\matr{H}$ consists of $l$ classifier predictions for $m$ data rows for the validation dataset.
Prediction speed of creating the matrix $\matr{H}$ depends on the choice of base classifier, number of data in the validation dataset ($m$), number of features ($d$), number of classes (i.e. binary), and data characteristics.
So, this study only investigates the computational complexity (see Table~\ref{ChapterEnsembleSD-Table-SDWEC-TrainingTimes}) and execution time (see Figure~\ref{figure-SDWEC-training-time}) of the proposed SDWEC training method (see Algorithm~\ref{code-SDWEC-Pseudo}) for the ensemble weight finding.
Note that, computational complexity of the SDWEC training only depends on number of data in validation set ($m$), number of classifiers ($l$), and number of algorithm iteration ($k$) (see table~\ref{ChapterEnsembleSD-Table-ComputationalComplexity}).  

    \item \textit{Applying ensemble on real-world data (i.e. test dataset)}:
Prediction time of SDWEC for test data (or unseen real-world data) depends on base classifiers prediction speed and number of base classifiers selected by SDWEC method (Algorithm~\ref{code-SDWEC-Pseudo}).
As the weights ($w$) of ensemble becomes more sparse (fewer non-zero elements in the solution $w$) fewer base classifiers are used in testing phase.
Thus, execution time of the testing time decreases as the weights become sparser independent of the employed base classifier.
In this study, weak decision tree classifier is used as a base classifier since it is fast in training and prediction; thus, testing time of the SDWEC mostly depends on the sparsity of the obtained ensemble weights.

\end{enumerate}

\subsection{Experimental results: sparsity}
\label{section-Results-Sparsity}

\begin{figure*}
\centering

        \includegraphics[width=0.85\textwidth,keepaspectratio]{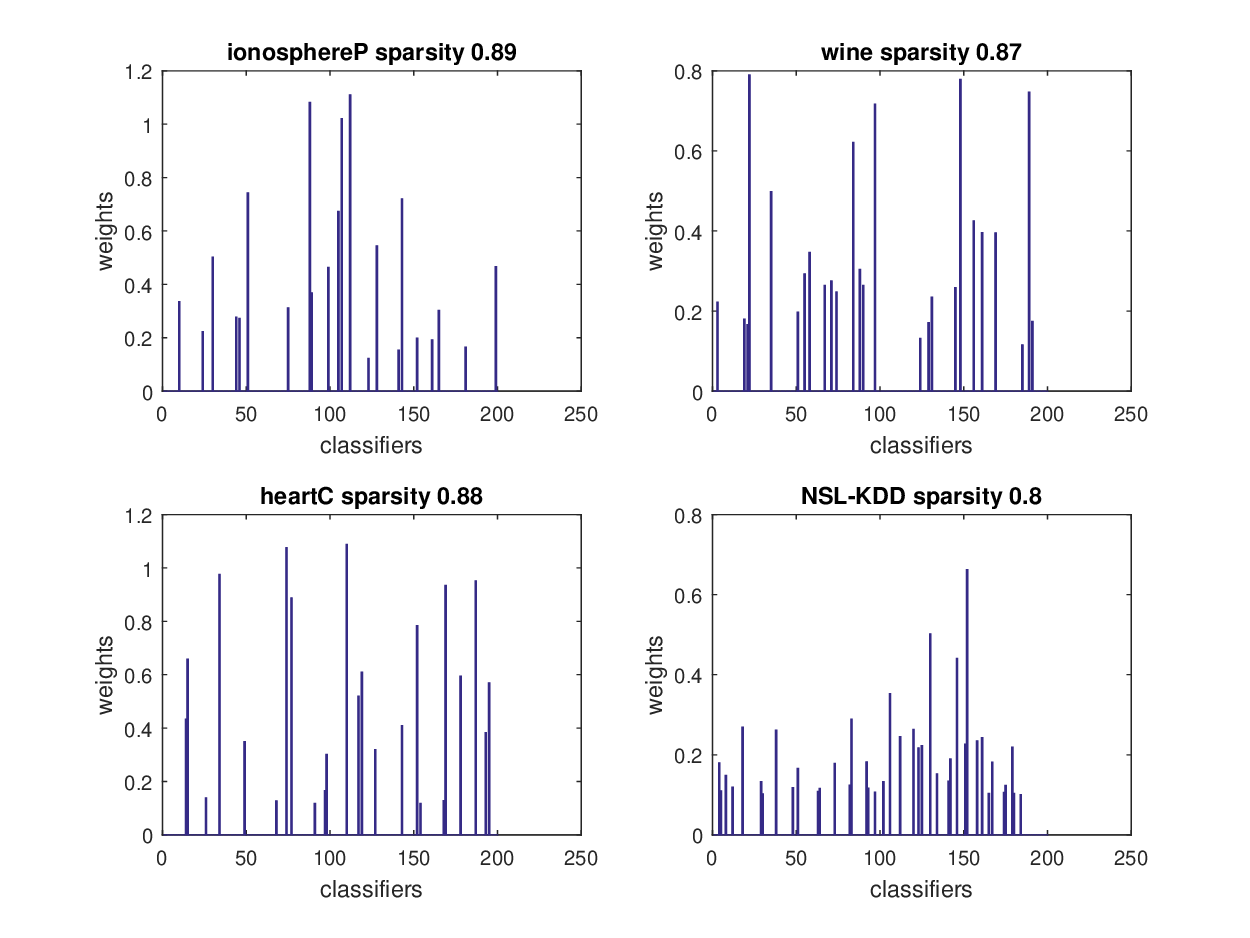}
        \vspace*{-5 mm}
        \caption[Datasets and their sparsity levels]{4 Datasets and their sparsity levels ($\lambda=1,\beta=10,\gamma=20,\epsilon=0.1$). }
        \label{figure-datasets-sparsity-levels}
         
\end{figure*}

The principle of parsimony (sparsity) states that simple explanation should be preferred to complicated ones \cite{Bach2012Optimization}.
Sparsity mostly used for feature selection in machine learning.
In our study, principle of sparsity is used for selecting subset of classifiers among weak classifiers.
During experiments, sparsity definition given in equation~\ref{eq-sparsity-definition} is used where $\mathscr{S}(w) = 0$ corresponds to least sparse solution while solution becomes more sparse as $\mathscr{S}(w)$ gets closer to $1$.
According to dataset and hyper-parameters used, SDWEC achieves different sparsity levels.
When SDWEC applied to 11 different datasets, sparsity levels between 0.80 and 0.88 has been achieved (Figure~\ref{figure-datasets-sparsity-levels}).
This means that among 200 weak classifiers, 24 classifiers (sparsity level of 0.88) to 40 classifiers (sparsity level of 0.80)  are used in ensembles.

\begin{align} 
\label{eq-sparsity-definition}
\begin{split}
    \mathscr{S}(w) = 1 - \frac{1}{l}||w||_0 
\end{split}
\end{align}
where
\begin{align} 
\label{eq-l0-norm-definition}
\begin{split}
    ||w||_0 = \#(r|w_r \neq 0), ~~~ (1 \leqslant r \leqslant l) 
\end{split}
\end{align}


Here, $||w||_0$ is the $L_0$-norm of a vector $w$.
Mathematically speaking, $L_0$-norm is not a proper norm since it is not absolutely homogeneous while it satisfies other norm properties.
In practice, $L_0$-norm is a cardinality function which has its definition in the form of $L_p$-norm for counting the number of non-zero elements in a given vector.

\begin{figure}[H]
\centering

        \includegraphics[width=\columnwidth,keepaspectratio]{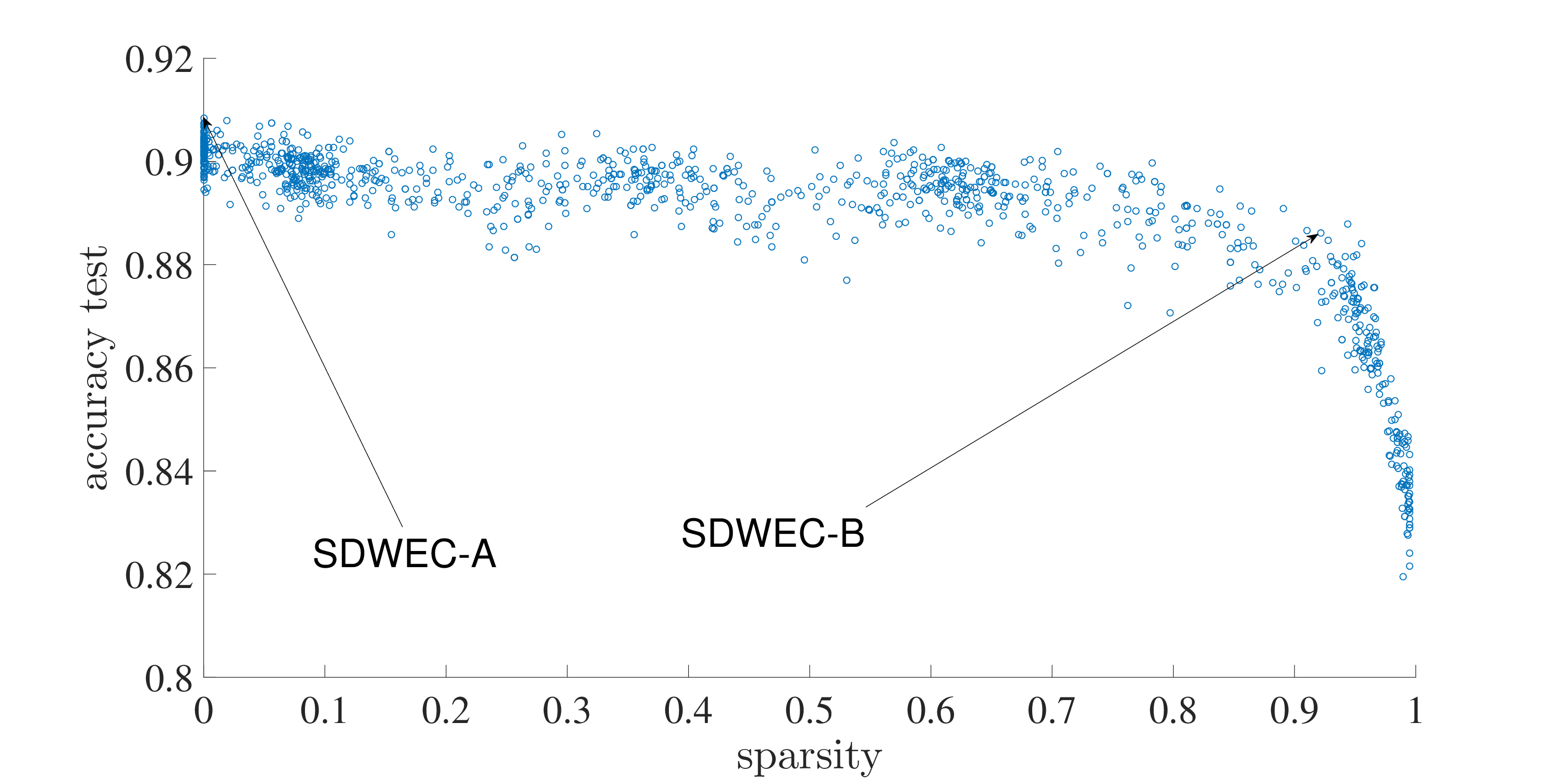}
        \vspace*{-6 mm}
        \caption[Sparsity vs accuracy of SDWEC.]{\textbf{Sparsity vs accuracy of SDWEC}. 
        The sparsity and accuracy values come from the mean of 11 datasets.
        Corresponding values can be seen in Table~\ref{table-comparison}. }
        \label{figure-sparsity-vs-accuracy-test}
 
\end{figure}

Two different results with different sparsity values (A and B), chosen from  Figure~\ref{figure-sparsity-vs-accuracy-test} have been provided in Table~\ref{table-comparison}.
SDWEC-A has no sparsity, all 200 base classifiers  have been used in ensemble; thus, it has superior performance at the cost of testing time.
SDWEC-A has best accuracy values in 4 out of 10 datasets and it is very close to top performing ones in others.
QFWEC is only slightly more accurate in 4 datasets comparing to SDWEC-A while SDWEC-A is only slightly more accurate comparing to QFWEC in other 4 datasets.
SDWEC provides similar accuracies with the best performing method (QFWEC) since both QFWEC and SDWEC-A use all base classifiers.
SDWEC-B has 0.90 sparsity, 20 of 200 base classifiers have been used in ensemble; nonetheless, it has best accuracy values in 2 out of 10 datasets.
Besides, its accuracy values are marginally lower (about 2\%) but its testing time is significantly better (90\%) than the other approaches.
Testing time of the methods in Table~\ref{table-comparison} is defined as $(1 - \mathscr{S}(.)) \mathscr{T}(l)$ where $\mathscr{S}(.)$ is the sparsity provided by the ensemble method (see equation~\ref{eq-sparsity-definition}) and $\mathscr{T}(l)$ is the testing time for all base classifiers. 
SDWEC-B has $10$ times faster testing time comparing to QFWEC since $\mathscr{S}(.)$ is $0$ for QFWEC and $0.9$ for SDWEC-B. 

\begin{table}[H]

\centering 
    \caption[Comparison of accuracies]{ \textbf{Comparison of accuracies (sparsity values are given in parentheses)}} 
    \label{table-comparison}

\setlength{\tabcolsep}{0.5em}

\begin{adjustbox}{width=\columnwidth,keepaspectratio}
\begin{tabular}{*{7}{r}}

\toprule
   Datasets  &   QFWEC  & SDWEC-A & SDWEC-B &   WMV   & bagging  & singleC  \\ 
\midrule
     breast   & 0.9736  & \textbf{0.9737} (0)  & 0.9532 (0.90) & 0.9355  & 0.9722  & 0.9400   \\
     heartC   & 0.8085  & 0.8186 (0) & \textbf{0.8279} (0.90) & 0.8118  & 0.8118  & 0.7268   \\
 ionosphere   & 0.9344  & 0.9371 (0) & \textbf{0.9427} (0.92) & 0.9371  & 0.9342  & 0.8799   \\
     sonarP   & 0.8088  & \textbf{0.8136} (0)  & 0.8126 (0.88) & 0.7893  & 0.8088  & 0.7367   \\
   vehicleP   & \textbf{0.9788} & 0.9693 (0)  & 0.9539 (0.91) & 0.9681  & 0.9670  & 0.9634   \\
      voteP   & 0.9576  & \textbf{0.9703} (0) & 0.9525 (0.84) & 0.8509  & 0.9703  & 0.9533   \\
   waveform   & \textbf{0.8812}  & 0.8652 (0)  & 0.8600 (0.93) & 0.8634  & 0.8620  & 0.8220   \\
      wdbcP   & \textbf{0.9595}  & 0.9507 (0)  & 0.9418 (0.88) & 0.9489  & 0.9507  & 0.9138   \\
       wine   & \textbf{0.9722}  & 0.9722 (0)  & 0.9605 (0.89) & 0.7514  & 0.9719  & 0.9500   \\
      wpbcP   & 0.7989  & \textbf{0.8036} (0)  & 0.7477 (0.91) & 0.7850  & 0.7750  & 0.6911   \\
      NSL-KDD   & 0.9828  & 0.9766 (0)  & 0.9849 (0.88) & 0.9610  & 0.9613  & \textbf{0.9976}  \\
\midrule
SDWEC-A  & \multicolumn{6}{l}{$\lambda=0.1 \textrm{ } \beta=35 \textrm{ } \gamma=5 \textrm{ } \epsilon=0.1 $ , Mean sparsity 0.00} \\ 
SDWEC-B  & \multicolumn{6}{l}{$\lambda=10   \textrm{ } \beta=15 \textrm{ } \gamma=15 \textrm{ } \epsilon=1.0 $, Mean sparsity 0.90  } \\ 

\bottomrule

\end{tabular}

\end{adjustbox} 
\scriptsize
     
\end{table}



\subsection{Computational Complexity Analysis}
\label{section-MaterialAndMethods-ComputationalComplexity}

In this section, computational complexity of SDWEC (Algorithm~\ref{code-SDWEC-Pseudo}) has been analyzed. 
First, computational complexity of every pseudo-code line in Algorithm~\ref{code-SDWEC-Pseudo} is given in table~\ref{ChapterEnsembleSD-Table-ComputationalComplexity} and then overall computational complexity is determined.
In Table~\ref{ChapterEnsembleSD-Table-ComputationalComplexity}, $m$ stands for the number of data in the validation dataset, $l$ stands for the number of base classifiers, and $k$ stands for the iteration count.

\begin{table}[H]
\centering
\caption{Computational complexity of SDWEC}
\label{ChapterEnsembleSD-Table-ComputationalComplexity}

\begin{adjustbox}{width=\columnwidth,max totalheight=0.8\textheight,keepaspectratio}

\begin{tabular}{p{0.05\linewidth}p{0.45\linewidth}p{0.25\linewidth}p{0.35\linewidth}}

\toprule
\textbf{Line} & \textbf{Code in Alg~\ref{code-SDWEC-Pseudo}} & \textbf{Complexity} & \textbf{Notes} \\

\midrule
    6   & $\hat{w} \gets w$    &   $O(l)$    &     \\
\midrule
    7   & $\gamma_r \gets \frac{\gamma}{|\hat{w}| + \epsilon} $    &    $O(l)$    &   \\
\midrule
    8   & construct $\matr{S}$ as diagonal form of $S_s$    &   $O(ml)$     & $ \matr{S} \leftarrow S_s $ sparse diagonal matrix $ (m \times m) $ (Eq~\ref{eq-Ss})  \\
\midrule
    9   & $v_B$    &   $O(l)$    &     \\
    9   & $\matr{C}$    &   $O(l)$    &  $\matr{C}$  sparse diagonal matrix  \\
\midrule
    10  & $\matr{S}\matr{H}$    &   $O(ml)$    &   $ \underset{m\times l}{\matr{X}} = \underset{m\times m}{\matr{S}} \times  \underset{m\times l}{\matr{H}} $   \\
    10  & $\matr{X}^{\intercal} \matr{X} $  &  $O(l^3)$  &  $ \matr{X}^{\intercal} : O(l^2) $, \newline  $ \matr{X}^{\intercal} \matr{X} : O(l^3) $   \\
    10  & $\matr{M} \gets \frac{2\lambda}{m}[\matr{X}^{\intercal} \matr{X}] + \frac{2(\matr{C}+\matr{I})}{l}$  &  $O(l^3+l^2)$  &  \\
\midrule
    11  & $\matr{X}^{\intercal}y$  &  $O(l^2)$  & \\
    11  & $\frac{2 \hat{w} - V_B}{l}$  &  $O(l)$  &  \\
    11  & $b \gets \frac{2\lambda}{m}\matr{X}^{\intercal}y + \frac{2 \hat{w} - V_B}{l}$  &  $O(l^2+l) $  & \\
\midrule
    12  & solve $ \matr{M}w = b $  &  $O(l^3)$  &  Cholesky solver  \\

\midrule
      13 & $ w = max(w,0)
$  &  $O(l)$  & \\
\bottomrule
\end{tabular}
\end{adjustbox} 
\small {$\matr{M}$ is dense, symmetric, real, and positive definite. \\ 
Cholesky solver is used to solve $\matr{M}w = b$, $O(\frac{2}{3}l^3)$. }
 
\end{table}

\begin{figure}[H]
\centering

        \includegraphics[width=\columnwidth,keepaspectratio]{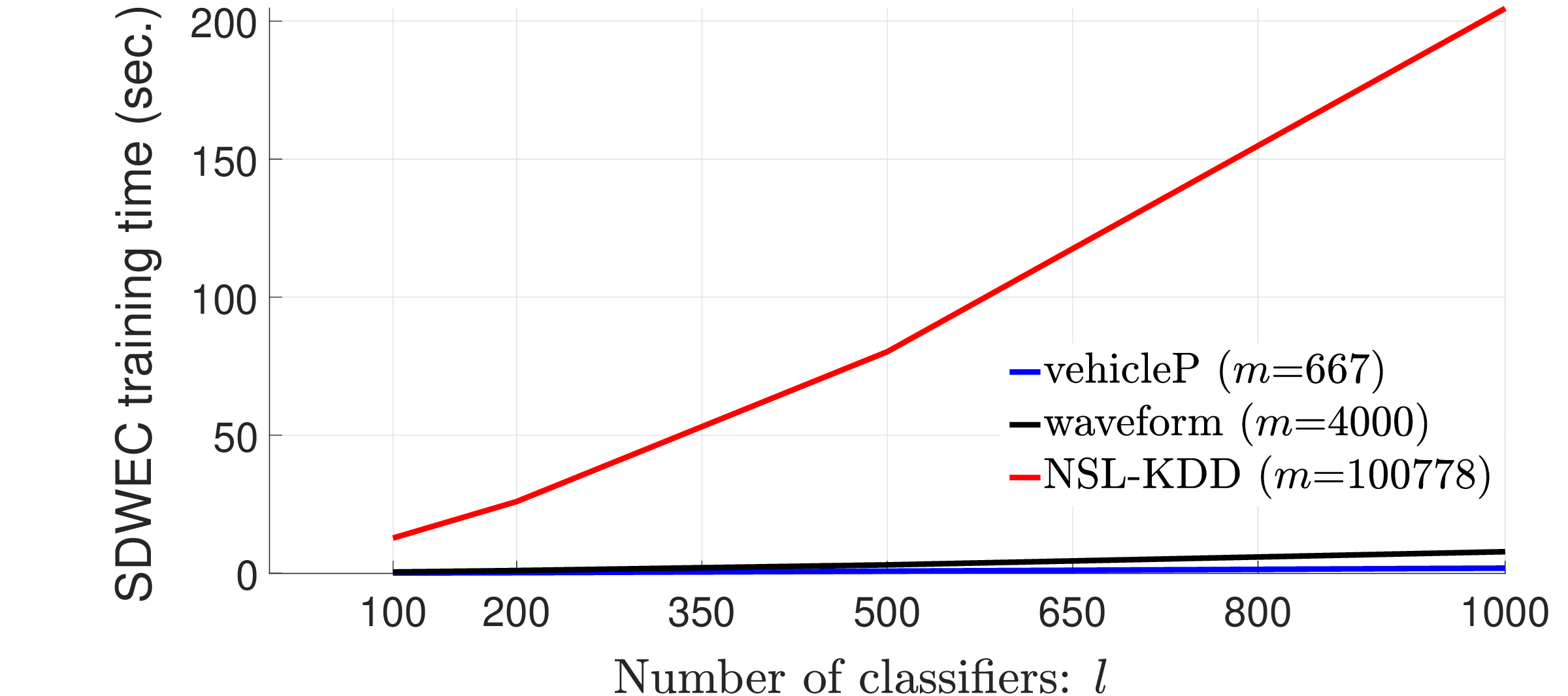}
         \vspace*{-5 mm}
        \caption{Number of classifier ($l$) versus SDWEC training time (see Table~\ref{ChapterEnsembleSD-Table-SDWEC-TrainingTimes}).}
        \label{figure-SDWEC-training-time}
 
\end{figure}

Computational complexity inside the for loop is $ O(ml) + C_1 O(l^3) + C_2 O(l^2) + C_3 O(l)$.
Since $ l \ll m $, dominant term is $ O(ml) $  for the $\matr{S}\matr{H}$ multiplication in line 10 of the Algorithm~\ref{code-SDWEC-Pseudo}, where $\matr{S}$ is a diagonal matrix.
Our iteration count is $k$, then final computational complexity of SDWEC is $O(kml)$, that is linear in $k$, $m$, and $l$ (see Table~\ref{ChapterEnsembleSD-Table-SDWEC-TrainingTimes} and Figure~\ref{figure-SDWEC-training-time}). 
This computational complexity analysis shows the computational efficiency of the proposed numerical minimization.

Table~\ref{ChapterEnsembleSD-Table-SDWEC-TrainingTimes} shows training time (weight finding) of SDWEC on various datasets.
Note that, $\matr{H}$ is an input to the Algorithm~\ref{code-SDWEC-Pseudo} and calculated as a prior step; thus, training times given in Table~\ref{ChapterEnsembleSD-Table-SDWEC-TrainingTimes} only corresponds to SDWEC training.
In this experiment, execution time is only dependent on number of rows ($m$) and number of classifiers ($l$) since fixed iteration count is used ($k = 25$).
In training set, NSL-KDD dataset (100778 instances) has 25 times more instances than waveform dataset (4000 instances).
And training time of NSL-KDD (25.95) is about 25 times of waveform (0.96).
In Figure~\ref{figure-SDWEC-training-time}, SDWEC training times are shown for 3 datasets with different number of data ($m$), different number of classifiers ($l$), and for fixed iteration count. 
As seen in Table~\ref{ChapterEnsembleSD-Table-SDWEC-TrainingTimes} and Figure~\ref{figure-SDWEC-training-time}, practical execution times are in alignment with theoretical computational complexity analysis. 
Slight differences between theoretical analysis and actual execution times are due to implementation issues and caching in CPU architectures.

\begin{table}[H]
\centering
\caption{SDWEC training time on various datasets, }
\label{ChapterEnsembleSD-Table-SDWEC-TrainingTimes}
\begin{adjustbox}{width=\columnwidth,keepaspectratio}
\begin{tabular}{lllllll}
\toprule
\textbf{Dataset}       & \textbf{Rows ($m$)}   & \multicolumn{4}{l}{\textbf{Time (sec.) $l$ classifier count }}  \\
              &        & $l=100$ & $l=200$ & $l=500$ & $l=1000$ \\
\midrule
breast-cancer & 547    & 0.05  & 0.10  & 0.48    & 1.63   \\
ionosphereP   & 280    & 0.04  & 0.07  & 0.31    & 1.01   \\
wpbcP         & 155    & 0.03  & 0.06  & 0.26    & 0.89   \\
wdbcP         & 456    & 0.05  & 0.09  & 0.44    & 1.34   \\
wine          & 143    & 0.03  & 0.05  & 0.23    & 0.91   \\
waveform      & 4000   & 0.43  & 0.96  & 3.01    & 7.78   \\
voteP         & 186    & 0.03  & 0.07  & 0.24    & 0.97   \\
vehicleP      & 667    & 0.06  & 0.18  & 0.73    & 1.83   \\
sonarP        & 167    & 0.03  & 0.06  & 0.23    & 0.83   \\
heartC        & 239    & 0.03  & 0.07  & 0.25    & 1.02   \\
NSL-KDD       & 100778 & 12.73 & 25.95 & 80.23   & 204.59 \\

\bottomrule
\end{tabular}
\end{adjustbox}
\end{table}


\section{Conclusion}
\label{section-Conclusion}
In this article, a novel sparsity driven ensemble classifier method, SDWEC, has been presented.
An efficient and accurate solution for original cost function (hard to minimize, non-convex, and non-differentiable) has been developed.
A novel convex relaxation technique for \textit{sign} function, and a novel adaptive \textit{log-sum-exp} approximation for the approximation of \textit{max} function that reduces numerical overflows are proposed.
Computational complexity of SDWEC has been investigated theoretically and experimentally.
SDWEC training has a linear computational complexity in number of classifier used ($l$), number of instances in the validation dataset ($m$), and  number of algorithm iterations ($k$).
SDWEC has been compared with other ensemble methods in well-known UCI and NSL-KDD datasets.
According to the experiments, SDWEC decreases number of classifiers used in ensemble without significant loss of accuracy.
By tuning parameters of SDWEC, a more sparse ensemble --thus, better testing time-- can be obtained with a small decrease in accuracy.



\section*{Appendix}

Optimality conditions can be used to show strict convexity since equation~\ref{eq-cost-function-after-taylor-new-regularization-term-matrix-vector-form} is in matrix-vector form and differentiable.

In equation~\ref{eq-cost-function-system-of-linear-equations}, first derivative of equation~\ref{eq-cost-function-after-taylor-new-regularization-term-matrix-vector-form} is equalized to zero and a close-form solution is obtained as a linear system so first order optimality condition is satisfied.

Let $\matr{G}$ be a second derivative (Hessian) of the cost function $J^{(n)}(w)$ given in the equation~\ref{eq-cost-function-after-taylor-new-regularization-term-matrix-vector-form}.
A symmetric matrix $\matr{G} \in \mathbb{R}^{l x l}$ is called positive definite (thus $J^{(n)}(w)$ is strictly convex), denoted by $\matr{G} \succ 0$, if $x^{\intercal} \matr{G} x > 0$ for every $x \in \mathbb{R}^{l}$ with $x \neq 0$.
Lets take the second derivative of the convex-relaxed cost function $J^{(n)}(w)$ given in equation~\ref{eq-cost-function-after-taylor-new-regularization-term-matrix-vector-form}:

\begin{align} 
\label{eq-matrix-vector-form-cost-function-second-derivative}
\begin{split}
   \frac{\partial^2{J^{(n)}(w)}}{\partial{w}^2} = \frac{2\lambda}{m}(\matr{S}\matr{H})^{\intercal}(\matr{S}\matr{H}) + \frac{2}{l}\matr{C} = \matr{G}
\end{split}
\end{align}

Lets show that $x^{\intercal} \matr{G} x > 0$ for all non-zero $x$:

\begin{align} 
\label{eq-matrix-vector-form-cost-function-second-derivative-test-step1}
\begin{split}
   x^{\intercal}(\frac{2\lambda}{m}(\matr{S}\matr{H})^{\intercal}(\matr{S}\matr{H}) + \frac{2}{l}\matr{C})x > 0
\end{split}
\end{align}

If we distribute $x^{\intercal}$ and $x$ from left and right:

\begin{align} 
\label{eq-matrix-vector-form-cost-function-second-derivative-test-step2}
\begin{split}
   \frac{2\lambda}{m}x^{\intercal}(\matr{S}\matr{H})^{\intercal}(\matr{S}\matr{H})x + \frac{2}{l}x^{\intercal} \matr{C} x > 0
\end{split}
\end{align}

Since $\lambda$, $m$, and $l$ are all positive we just need to show that $x^{\intercal}(\matr{S}\matr{H})^{\intercal}(\matr{S}\matr{H})x > 0$ and $x^{\intercal} \matr{C} x > 0$:

\begin{align} 
\label{eq-matrix-vector-form-cost-function-second-derivative-test-step3}
\begin{split}
   x^{\intercal}(\matr{S}\matr{H})^{\intercal}(\matr{S}\matr{H})x > 0 \rightarrow   
     (\matr{S}\matr{H}x)^{\intercal}(\matr{S}\matr{H}x) > 0
\end{split}
\end{align}

Lets define $z$ as $z = \matr{S}\matr{H}x$, then $z^{\intercal}z > 0$ since $\matr{S}$ is a diagonal matrix with all positive elements (see equation~\ref{eq-Ss}), $\matr{H}$ contains non-zero elements $\{-1, 1\}$, and $x$ is non-zero vector.

$\matr{C}$ is a diagonal matrix with diagonal elements $C_r$, $1 \leq r \leq l$, which are defined as below (from second order Taylor approximation):

\begin{align} 
\label{eq-matrix-vector-form-cost-function-second-derivative-test-step4}
\begin{split}
   C_r = \frac{\gamma_r(4u^2_r + 8u^3_r + 4u^4_r + \beta u_r + 2\beta u^3_r + \beta u^5_r)}{2(u^3_r + u^2_r + u_r + 1)^2}
\end{split}
\end{align}
where $u_r = e^{\hat{w}_r \gamma_r}$.
Here, $\beta$ is a positive constant, $\gamma_r = \gamma (|\hat{w}_r| +\epsilon)^{-1}$ is always positive since $\gamma > 0$, and $u_r$ is always positive since $\hat{w}_r \gamma_r \geq 0$.
Thus, $x^{\intercal} \matr{C} x > 0$ is satisfied since $C_r$ is always positive.

Therefore, both first order optimality conditions and second order optimality conditions are satisfied which shows that cost function $J^{(n)}(w)$ given in equation~\ref{eq-cost-function-after-taylor-new-regularization-term-matrix-vector-form} is strictly convex.











\end{multicols}
\end{document}